# Robust Outlier Detection Technique in Data Mining: A Univariate Approach

Singh Vijendra and Pathak Shivani
Faculty of Engineering and Technology
Mody Institute of Technology and Science
Lakshmangarh, Sikar, Rajasthan, India

**ABSTRACT**

Outliers are the points which are different from or inconsistent with the rest of the data. They can be novel, new, abnormal, unusual or noisy information. Outliers are sometimes more interesting than the majority of the data. The main challenges of outlier detection with the increasing complexity, size and variety of datasets, are how to catch similar outliers as a group, and how to evaluate the outliers. This paper describes an approach which uses Univariate outlier detection as a pre-processing step to detect the outlier and then applies K-means algorithm hence to analyse the effects of the outliers on the cluster analysis of dataset.

## 1. INTRODUCTION

Data mining, in general, deals with the discovery of hidden, non-trivial and interesting knowledge from different types of data. As the development of information technologies is taking place, the number of databases, as well as their dimension and complexity is growing rapidly. Hence there is a need of automated analysis of great amount of information. The analysis results which are then generated are used for making a decision by a human or program. Outlier detection is one of the basic problems of data mining. An outlier is an observation of the data that deviates from other observations so much that it arouses suspicions that it was generated by a different and unusual mechanism [6]. On the other hand, Inlier is defined as an observation that is explained by mechanism underlying probability density function. This function represents probability distribution of main part of data observations [2].

Outliers may be erroneous or real. Real outliers are observations whose actual values are very different from those observed for the rest of the data and violate plausible relationships among variables. Erroneous outliers are observations that are distorted due to misreporting errors in the data-collection process. Both types of outliers may exert undue influence on the results of statistical analysis, so they should be identified using reliable detection methods prior to performing data analysis [7].Outliers are sometimes found as a side-product of clustering algorithms. These techniques define outliers as points, which do not lie in any of the clusters formed. Thus, the techniques implicitly define outliers as the background noise in which the clusters are embedded. Another class of techniques are also there which defines outliers as points, which are neither a part of a cluster nor a part of the background noise; rather they are specifically points which behave very differently from the normal data [1].The problem of detecting outliers has been studied in the statistics community as well. In the approaches thus designed, user has to model the data points using a statistical distribution, and points are determined to be outliers depending on how they appear in relation to the postulated model. The main problem with these approaches is that in a number of situations, the user might not have enough knowledge about the underlying data distribution [5].

Sometimes outliers can often be considered as individuals or groups of clients exhibiting behaviour outside the range of what is considered normal. Outliers can be removed or considered separately in regression modelling to improve accuracy which can be considered as benefit of outliers. Identifying them prior to modelling and analysis is important [6] so that the analysis and the results might not get affected due to them. The regression modelling consists in finding a dependence of one random variable or a group of variables on another variable or a group of variables. In the context of outlier-based association method, outliers are observations markedly different from other points. When a group of points have some common characteristics, and these common characteristics are "outliers", these points are associated [4] with each other in some or the other way. Statistical tests depend on the distribution of the data whether or not the distribution parameters are known, the number of excepted outliers, the types of excepted outliers [3].

## 2. RELATED WORK

A survey of outlier detection methods was given by Hodge & Austin [8], focusing especially on those developed within the Computer Science community. Supervised outlier detection methods, are suitable for data whose characteristics do not change through time, they have training data with normal and abnormal data objects. There may be multiple normal and/or abnormal classes. Often, the classification problem is highly imbalanced. In semi-supervised recognition methods, the normal class is taught, and data points that do not resemble normal data are considered outliers. Unsupervised methods process data with no prior knowledge. Four categories of unsupervised outlier detection algorithms; (1) In a clustering-based method, like DBSCAN (a density-based algorithm for discovering clusters in large spatial databases) [9], outliers are by-products of the clustering process and will not be in any resulting cluster. (2) The density-based method of [10] uses a Local Outlier Factor (LOF) to find outliers. If the object is isolated with respect to the surrounding neighborhood, the outlier degree would be high, and vice versa. (3) The distribution-based method [3] defines, for instance, outliers to be those points $p$ such that at most 0.02% of points are within 0.13 $\sigma$ of $p$. (4) Distance-based outliers are those objects that do not have "enough" neighbours [11][12]. The problem of finding outliers can be solved by answering a nearest neighbour or range query centered at each object $O$.

Several mathematical methods can also be applied to outlier detection. Principal component analysis (PCA) can be used to detect outliers. PCA computes orthonormal vectors that provide a basis (scores) for the input data. Then principal components are sorted in order of decreasing "significance" or strength. The size of the data can be reduced by eliminating the weaker components which are with low variance [13]. The convex hull method finds outliers by peeling off the outer layers of convex hulls [14]. Data points on shallow layers are likely to be outliers.

## 3. UNIVARIATE OUTLIER ANALYSIS

Univariate outliers are the cases that have an unusual value for a single variable. One way to identify univariate outliers is to convert all of the scores for a variable to standard scores. If the sample size is small (80 or fewer cases), a case is an outlier if its standard score is ±2.5 or beyond. If the sample size is larger than 80 cases, a case is an outlier if its standard score is ±3.0 or beyond. This method applies to interval level variables, and to ordinal level variables that are treated as metric.

## 4. K-MEANS CLUSTERING

In data mining cluster analysis can be done by various methods. One such method is K-means algorithm which partition n-observations into k-clusters in which each observation belongs to the cluster with the nearest mean. K-means clustering operates on actual observations and a single level of clusters is created. K-means clustering is often more suitable than hierarchical clustering for large amounts of data. Each observation in your data is treated as an object having a location in space. Partitions are formed based upon the fact that objects within each cluster are as close to each other as possible, and as far from objects in other clusters as possible. We can choose from five different distance measures, depending on the kind of data we are clustering. Each cluster in the partition is defined by two things, its member objects and centroid (the point to which the sum of distances from all objects in that cluster is minimized), or centre. Cluster centroids are computed differently for each distance measure, to minimize the sum with respect to the measure that you specify. An iterative algorithm is used, that minimizes the sum of distances from each object to its cluster centroid, over all clusters. This algorithm keeps on moving the objects between clusters until the sum cannot be decreased further giving a set of clusters that are as compact and well-separated as possible.

## 5. PROPOSED APPROACH

As we have reviewed several different ways of detecting outliers we here propose a method which is a combination of two approaches, statistical and clustering. Firstly we apply an Univariate outlier detection in SPSS and then k-means algorithm to group the data into clusters. Following are the steps of the algorithm in detail:

Step 1: Choose a dataset on which outlier detection is to be performed.

Step 2: Apply Univariate Outlier Detection in SPSS to do the pre-processing of data before applying cluster analysis.

Step 3: If Outlier is detected based upon the criterion

- If the sample size is small (80 or fewer cases), a case is an outlier if its standard score is ±2.5 or beyond.
- If the sample size is larger than 80 cases, a case is an outlier if its standard score is ±3.0 or beyond

Then run the k-means (clustering) algorithm, for the dataset with and without the tuple having outlier, using replicates in order to select proper centroids so as to overcome the problem of local minima.

Step 4: Compare the results for the sum of point-to-centroid distances. The main goal of k-means algorithm is to find the clusters in such a way so as to minimize the sum of point to centroid distance.

Step 5: If Sum (without outlier) < Sum (with outlier), then remove the tuple having outlier permanently from the dataset.

## 6. EXPERIMENTAL RESULTS

We used a hypothetical data which has been made by introducing some outlier values in the well known fisher iris data, having four features namely sepal length, sepal width, petal length, petal width and 150 tuples. After making the dataset we applied Univariate outlier detection on it using SPSS, a widely used Statistical Tool. Results of the univariate outlier analysis are shown in Figure.1 case as there are more than 80 cases. We arrange the z-scores in ascending and descending order and hence search for any value lesser than -3 or greater than +3. Here we get a value greater than +3 when we arranged the z-scores in descending order, value shown in Figure.4.

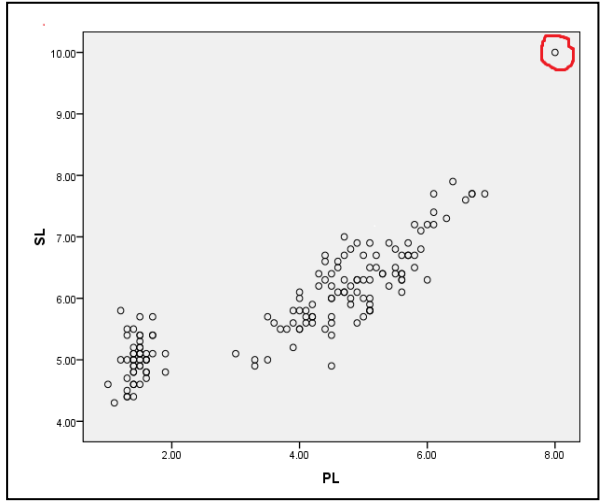

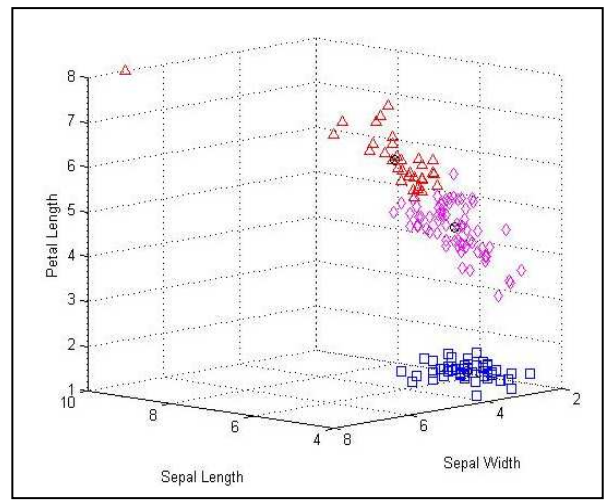

Figure .4. Values of z-score, showing value > +3

Figure.1.Plot between sepal length and petal length in the presence of outlier

The plot shows the distribution of points between the two features of our data Sepal length and Petal length, and the highlighted point in red circle shows the outlier point. The descriptive univariate analysis applied on sepal length feature of the data (with outlier) gives following results, shown in Figure.2. whereas that of analysis of data (without outlier) is shown in Figure.3.

Now we must delete the outlier entry and save both the dataset i.e. with outlier entry and without outlier entry and run further the k-means algorithm to do the cluster analysis of the data and calculate the sum of points to centroid value in each case. After calculating both the values we will compare them to check whether the presence of outlier increases the sum or not, if it increases then it must be removed. Figure.5. and shows the clustering analysis of the dataset with outlier.

**Descriptive Statistics**

|  | N | Minimum | Maximum | Mean | Std. Deviation |
|---|---|---|---|---|---|
| SL | 151 | 4.30 | 10.00 | 5.8709 | .89193 |
| Valid N (listwise) | 151 | | | | |

Figure.2. Descriptive univariate analysis of the data (with outlier)

**Descriptive Statistics**

|  | N | Minimum | Maximum | Mean | Std. Deviation |
|---|---|---|---|---|---|
| SL | 150 | 4.30 | 7.90 | 5.8433 | .82807 |
| Valid N (listwise) | 150 | | | | |

Figure.3. Descriptive univariate analysis of data (without outlier)

Figure .5. Cluster analysis with outlier

We can easily observe here that the mean calculated in Figure.2. is 5.8709 and the maximum value is 10 i.e. much greater than the mean whereas in Figure.3.its nearer to the mean but statistically we observe the z-scores of the sepal length calculated by the analysis, these scores must follow a range of -3 to +3 in our

Value of the sum of all points in the cluster to the centroid comes out to be 116.1295, shown in Figure.6. Now we will calculate the sum value for the clustering of data without outlier entries. Figure.7. shows the

Figure.6. Value of the sum clustering analysis of the dataset without outlier.

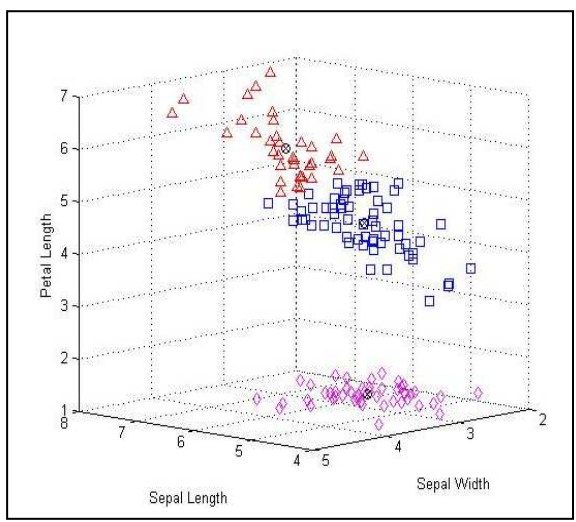

Figure .7. Cluster analysis without outlier

Figure.8. Value of the sum

Value of the sum of all points in the cluster to the centroid comes out to be 78.8514, shown in Figure.8. which is much lesser than the sum value of the dataset with outlier entries. Hence we will delete the outlier entry permanently from the dataset, because this entry is not at all useful and distorting our original dataset.

## 7. CONCLUSION

The conclusion of the whole paper lies in the fact that outliers are usually the unwanted entries which always affects the data in one or the other form and distorts the distribution of the data. Sometimes it becomes necessary to keep even the outlier entries because they play an important role in the data but in our case as we are achieving our main objective i.e. to minimize the value of the sum by choosing optimum centroids, we will delete the outlier entries.